\documentclass[11pt]{article}

\usepackage[preprint]{coling}

\usepackage{times}
\usepackage{latexsym}

\usepackage[T1]{fontenc}

\usepackage{subcaption}

\usepackage{enumitem}

\usepackage{microtype}

\usepackage{inconsolata}

\usepackage{graphicx}
\usepackage{multirow}

\usepackage{color}
\usepackage{xcolor}
\definecolor{lightyellow}{RGB}{255, 204, 99} 
\definecolor{lightred}{RGB}{255, 204, 203}
\definecolor{lightgreen}{RGB}{185, 224, 165}
\usepackage{booktabs}
\usepackage{amsmath} 
\usepackage{amssymb}
\usepackage{amsxtra}
\usepackage{fancyhdr}
\usepackage{hyperref}
\usepackage{graphicx}
\usepackage{longtable}        
\usepackage{tabularx}
\usepackage{soul}
\usepackage{arydshln}

\title{Fine-Tuning LLMs for Low-Resource Dialect Translation: The Case of Lebanese}

\author{Silvana Yakhni, Ali Chehab\\
        Electrical and Computer Engineering\\
        American University of Beirut\\
        \texttt{syy06@mail.aub.edu, chehab@aub.edu.lb}
}
 
\begin{document}
\maketitle

\begin{abstract}
This paper examines the effectiveness of Large Language Models (LLMs) in translating the low-resource Lebanese dialect, focusing on the impact of culturally authentic data versus larger translated datasets. We compare three fine-tuning approaches: Basic (\texttt{Instruct-MT}), contrastive (\texttt{Instruct-Cont}), and grammar-hint (\texttt{Instruct-Grammar}) tuning, using open-source Aya23 models. Experiments reveal that models fine-tuned on a smaller but culturally aware Lebanese dataset (LW) consistently outperform those trained on larger, non-native data. The best results were achieved through contrastive fine-tuning paired with contrastive prompting, which indicates the benefits of exposing translation models to bad examples. In addition, to ensure authentic evaluation, we introduce LebEval, a new benchmark derived from native Lebanese content, and compare it to the existing FLoRes benchmark. Our findings challenge the \textbf{"More Data is Better"} paradigm and emphasize the crucial role of cultural authenticity in dialectal translation. We made our datasets and code available at \href{https://github.com/sylvanayakhni/Lebanese-Dialect-Translation}{Github}.
\end{abstract}

\section{Introduction}
Machine translation of dialectal Arabic presents a unique challenge that differs significantly from Modern Standard Arabic, including its rich cultural context and the scarcity of linguistic resources. This paper specifically focuses on the Lebanese dialect, a prominent Arabic variant in the Levant region. Although Large Language Models (LLMs) such as ChatGPT, LLaMA\citep{Touvron2023LLaMAOA} and BLOOM\citep{Scao2022BLOOMA1} have shown promising results in Machine Translation (MT) tasks  \citep{Hendy2023HowGA}  \citep{Jiao2023IsCA}, their effectiveness in handling culturally embedded dialects remains largely unexplored. Figure 1 shows a failed attempt by GPT-4o to translate a famous Lebanese idiom, highlighting the challenges of dialectal translation and raising questions about how to leverage LLMs for translating low-resource dialects. 

Recent work reveals a significant gap in harnessing LLMs for Arabic dialectal MT. While existing studies have predominantly focused on evaluating LLMs through zero-shot and few-shot prompting  \citep{Khondaker2024BenchmarkingLO}\citep{kadaoui2023tarjamat} \citep{Abid2020TheSE}, prompt-based approaches are inherently constrained by the model's pre-existing knowledge \citep{shin2023promptengineeringfinetuning} and may fall short in handling the complex cultural undertones and region-specific idioms. Notably, finetuning LLMs using translation instructions has been extensively explored in MT research and has demonstrated promising results \citep{Li2023ElicitingTT}\citep{Mao2024TuningLW} \citep{Jiao2023ParroTTD}. However, its application to Arabic dialectal translation remains largely unexplored. 

\begin{figure}[t]
    \centering
    \includegraphics[width=1\linewidth]{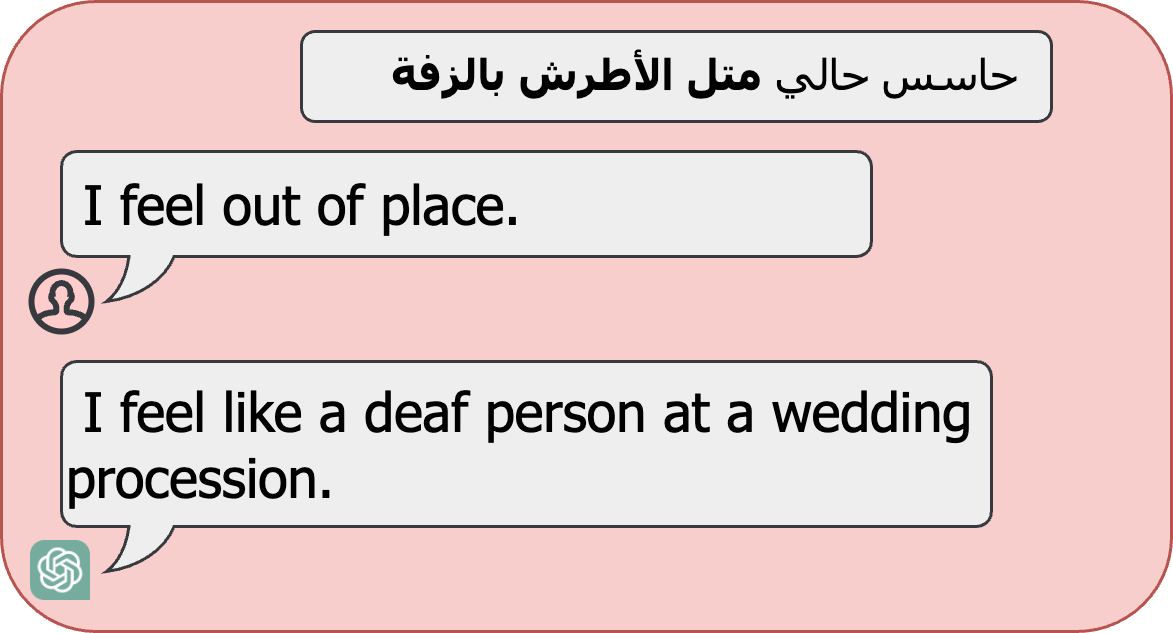}
    \caption{Example of the translation of a cultural Lebanese idiom by a human translator\includegraphics[height=1em]{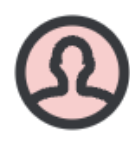} compared to GPT-4o\includegraphics[height=1em]{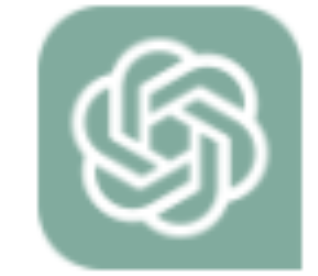}}
    \label{fig:enter-label}
\end{figure}

This study addresses the limitations in Arabic dialect translation by conducting a systematic comparison of fine-tuning and prompting techniques on the open-source Aya23-8B model \citep{aryabumi2024aya23openweight}. Our methodology encompasses four distinct approaches: Fine-tuning using 1) basic, 2) contrastive, and 3) grammatical-hint instructions. In addition, we explored the effect of 4) curriculum learning for grammar rules, and translation quality acquisition before translation. 

A key contribution of this work is the investigation of data quality in translating culturally rich dialectal content. We examined the impact of fine-tuning on the Lebanese culturally aware dataset: LanguageWave(LW) \citep{yakhni-chehab-2025-llms} compared to non-authentic translated data: MADAR\citep{Bouamor2018TheMA} and OpenSubtitles (OS)\citep{Krubiski2023MultiParallelCO}. Furthermore, we present LebEval, a novel evaluation dataset sourced from authentic Lebanese content, addressing the prevalent limitation of existing benchmarks that primarily rely on translated materials. 

Our experimental results reveal several key findings: 
\begin{itemize}
    \item Fine-tuning LLMs using culturally aware datasets yields superior results across all prompting techniques, emphasizing the critical role of data quality over quantity.
    \item Fine-tuning using contrastive instructions surpasses fine-tuning using basic instructions, particularly when paired with contrastive prompting, demonstrating the value of using translation errors in the learning process. 
    \item Curriculum learning strategies yielded limited performance gains, likely due to catastrophic forgetting, a known challenge in LLMs. 
    \item The use of authentic evaluation datasets is essential for accurately assessing the ability of LLMs to translate dialectal content, as it better reflects the complexities of real-world linguistic and cultural nuances.
    
\end{itemize} 

Through this research, we seek to establish a more robust framework for dialectal translation using LLMs, which preserves the richness of Lebanese cultural expressions and enhances the model's reasoning capabilities in handling complex linguistic patterns.

\section{Instruction Pool}

\begin{figure*}[h]
    \centering
    \includegraphics[width=1\linewidth]{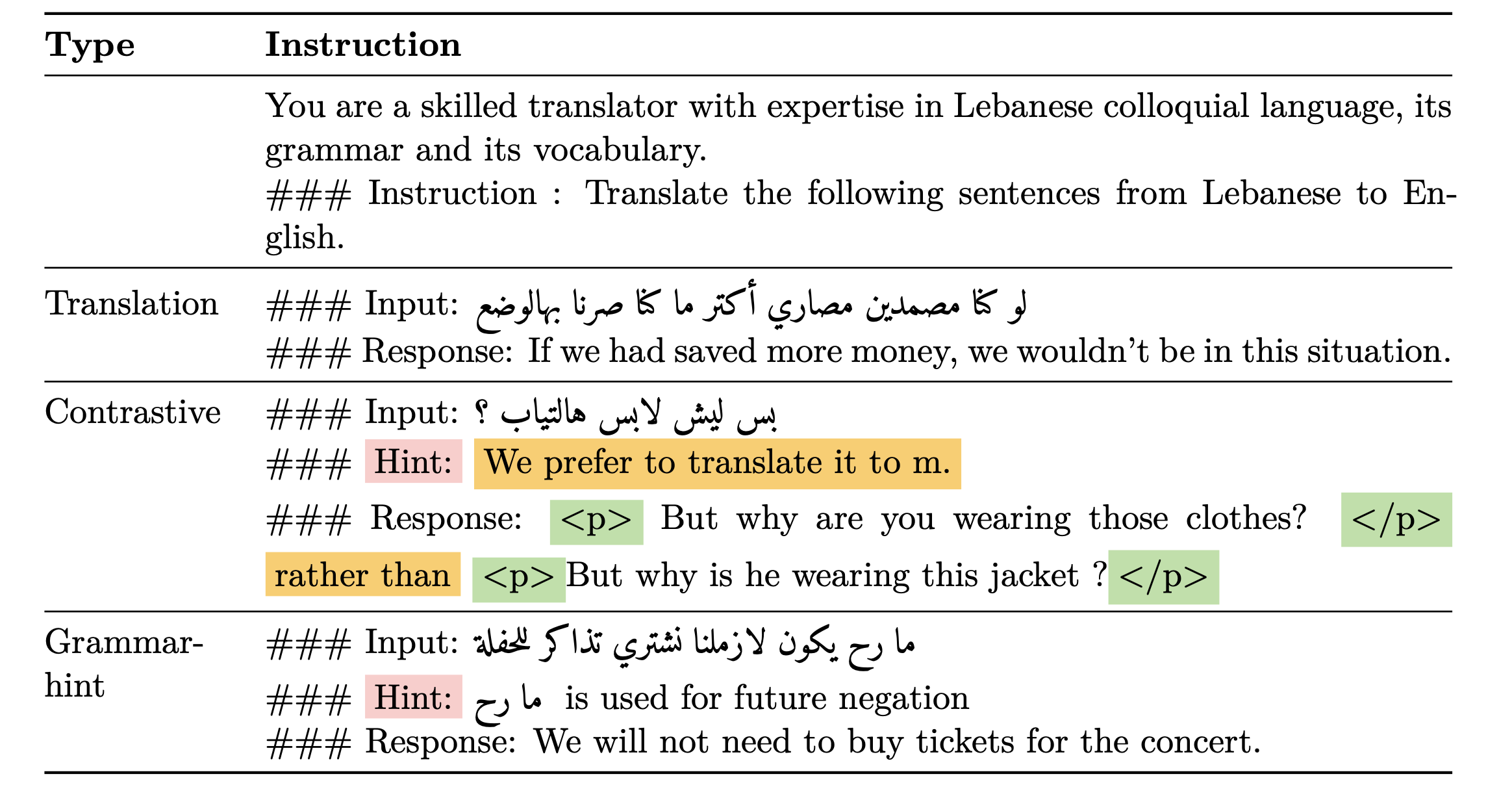}
    \caption{Translation Instructions Templates}
    \label{fig:enter-label}
\end{figure*}

LLMs are decoder-based models trained on next-word prediction referred to as "Causal Objective". Hence, Supervised Fine-Tuning (SFT) is used to train these models on parallel instruction data (prompt/answer) to produce desired outputs. In this section, we introduce three main types of instruction:  \textbf{1) Translation}, \textbf{2) Contrastive}, and \textbf{3) Grammar-hint}  instructions. The first type guarantees basic translation ability, while the last two regulate the LLM to develop a deeper understanding of different modes of translation failure. Figure 2 shows examples of each instruction.

\subsection{Translation Instruction}  
As traditional translation systems, we rely on bilingual sentence pairs to achieve the basic translation ability of LLMs. We follow the chat format adopted in \citep{alpaca} to transform bilingual sentence pairs into the instruction-following format. Figure 2 presents an example of translation instructions, which include a preamble and an instruction fixed for all tasks, usually establishing context, an “\#\#\#Input” with the source Lebanese sentence, and a “\#\#\# Response” with the target English sentence to be generated.

\subsection{Contrastive Instruction} 

By fine-tuning using contrastive instructions, we want LLMs to discern relative quality differences among translations. Achieving this objective requires ranking datasets. In our work, we identify two translations for each input sentence: a chosen/preferred translation and a rejected/undesirable translation as follows:
\begin{itemize}[left=0pt]
    \item Chosen answers are the golden translations.
    \item Rejected translations are generated from the base LLM Aya23-8b as suboptimal answers.
\end{itemize}

As presented in Figure 2, we construct the "\#\#\# Response" by concatenating two translations (e.g., joined by "<rather than>"), where the first translation represents the preferred choice. In addition, we include a "\#\#\#Hint" field to indicate our preference. The good and bad examples are separated using the \verb|<p>| delimiter. Essentially, the second translation serves as a negative sample within the sentence pair. 

\subsection{Grammar-Hint Instruction}
A potential limitation of contrastive instruction is that it indicates quality differences between translations without providing explicit guidance on how to improve them. To address this, we aim to enable LLMs to reason before translating by incorporating knowledge of vocabulary and grammatical rules specifically relevant to translation. For example, in Lebanese Arabic, the term "rah", if attached to a verb, is usually used to indicate a future tense. Teaching these rules can allow the model to better interpret and produce accurate translations. We achieve this by introducing a hint field in the training data, explicitly indicating the relevant grammatical or vocabulary rule, thereby encouraging reasoning prior to translation. Given time and resource constraints, we made an attempt to synthesize this dataset. 
\newline\newline\textcolor{red}{\textbf{Data Synthesis.  }} Given a grammatical Lebanese chapter with a set of rules accompanied by illustrative examples, we employed Claude 3.5 Sonnet\footnote{https://claude.ai} to generate relevant, coherent, and contextually rich translation examples. Figure 5 in Appendix B shows the process of synthesizing the Grammatical data along with the prompt used to instruct Claude. Figure 2 shows a sample of the resulting Grammatical-guided instruction.  
\newline\newline\textcolor{red}{\textbf{Why Claude 3.5 Sonnet rather than ChatGPT or Gemini?}} Claude 3.5 Sonnet was selected over alternatives such as ChatGPT or Gemini due to its demonstrated strength in generating descriptive and literary content. Additionally, its extended context length allowed us to process entire chapters from books in a single prompt.

\section{Experimental Setup}

\subsection{Training Data}
\textbf{Non-Native Data (NN). } Datasets for Lebanese-English translation are limited, with only a few available, such as Open Subtitles (OS)\citep{Krubiski2023MultiParallelCO}, which comprises 128K sentences of movie subtitles recently translated into Lebanese, and MADAR \citep{Bouamor2018TheMA}, a dataset of 12K travel-related feedback translated into Lebanese. However, these datasets share a critical limitation: they rely on translations from non-native sources, leading to a lack of cultural authenticity and contextual relevance. Together, these datasets amount to a total of 140K sentences, which we collectively refer to as Non-Native (NN) data. \newline\newline\textbf{Culturally-Aware Data (LW). } Recent research \citep{yakhni-chehab-2025-llms} introduced the Language Wave (LW) dataset, a culturally-aware Lebanese-English parallel dataset derived from a Lebanese podcast. The dataset consists of approximately 3K sentences extracted from 95 podcast episodes, which explore various aspects of Lebanese culture.\newline\newline\textbf{Lebanese Grammar Instruction Data (LGID). }  To create Grammatical instructions from Lebanese Grammar Chapters, we leveraged a Lebanese Grammar book titled "The Fundamentals of Lebanese Grammar" \citep{grammar_book}. The book comprises 32 Grammatical chapters, each providing a set of rules along with examples. Through the approach described in section 2.3, we compiled a dataset of 2,836 parallel Lebanese-English sentences, each annotated with a corresponding grammatical hint.  The collection process is illustrated in Figure 5 in Appendix B.

\subsection{Model Training}

\begin{figure}[h]
    \centering
    \includegraphics[width=1\linewidth]{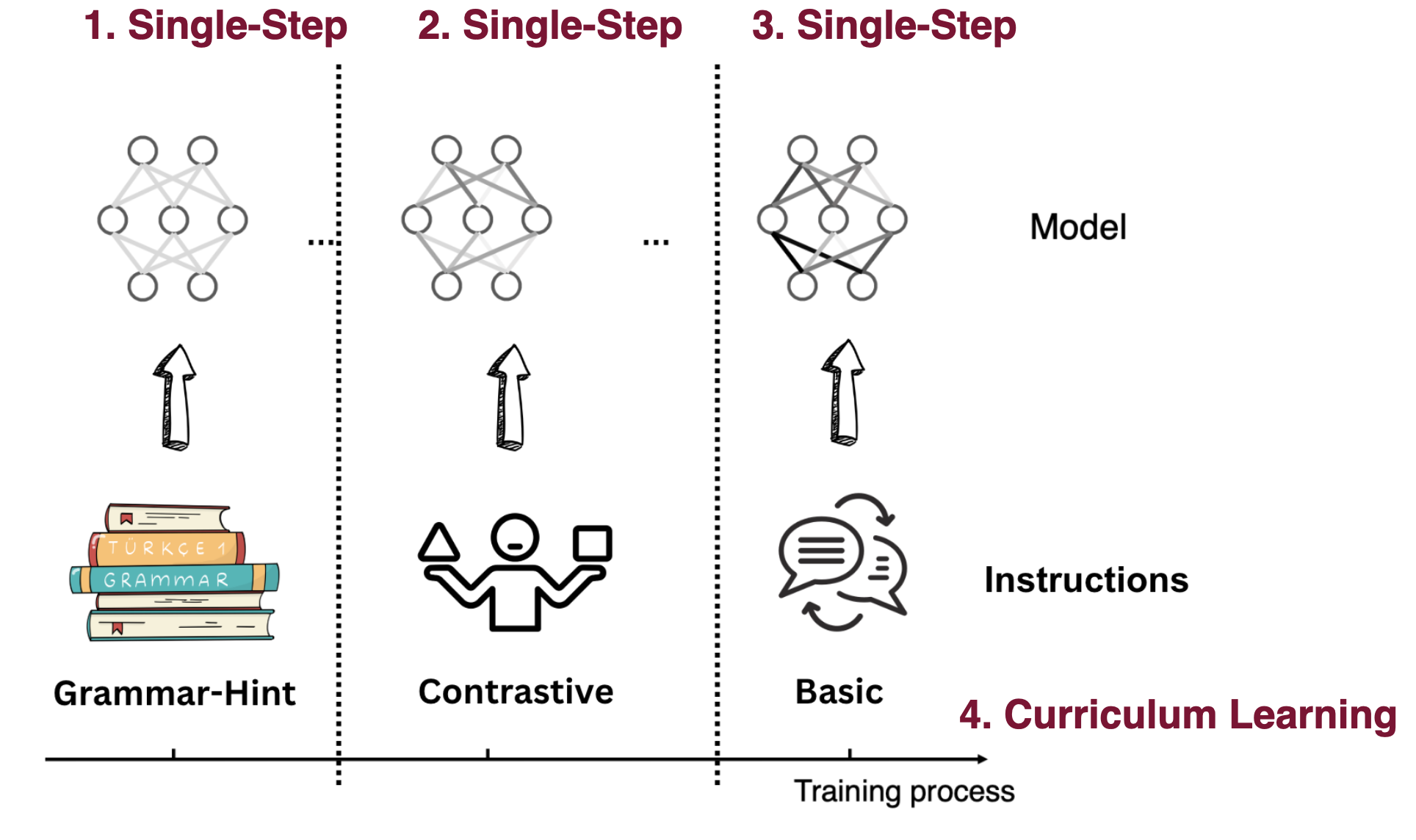}
    \caption{Illustration of four single-step and Curriculum Training Configurations}
    \label{fig:enter-label}
\end{figure}

We based our experiments on Aya23-8B model from Cohere AI \citep{aryabumi2024aya23openweight}, an open-source multilingual LLM developed with the help of native speakers to preserve cultural and linguistic authenticity. We investigated four training configurations, illustrated in Figure 3:

\begin{enumerate}
    \item \textbf{Single-Task Training}: Models were trained individually on specific instruction sets:
    \begin{itemize}
        \item \texttt{Instruct-MT}: fine-tuned on Machine translation instructions. 
        \item \texttt{Instruct-Cont}: fine-tuned on Contrastive instructions.
        \item \texttt{Instruct-Grammar}: fine-tuned on Grammar-based instructions. 
    \end{itemize} 
    \item \textbf{Curriculum Training 1}: arranges to fine-tune in a two-stage curriculum, first with Contrastive instructions, then with translation instructions.  This configuration validates if the performance could benefit from learning to distinguish between good and bad examples before translating. We refer to this curricula as \texttt{Cont+MT}.
    \item  \textbf{Curriculum Training 2}: start training on Grammar-Instructions, followed by Contrastive instructions, and finally with translation instructions. This curriculum tries to benefit from learning language rules (related to translation) before focusing on translation. We refer to this training as \texttt{Grammar+Cont+MT}. 
\end{enumerate}

Each training curriculum was applied to both the NN and LW datasets, resulting in two versions per curriculum:
\begin{itemize}
    \item \texttt{NN}-trained: Models trained on instructions derived from Non-Native data.
    \item \texttt{LW}-trained: Models trained on instructions derived from Culturally-Aware data. 

\end{itemize} 
In case of multi-step training curricula, the instruction datasets were split into two subsets: 50\% allocated to generate translation instructions and 50\% reserved for creating contrastive instructions. This strategy balanced the focus on both instruction fidelity and contrastive learning.\newline

For both basic and contrastive fine-tuning,  we used Qlora for efficient fine-tuning, with a Lora rank 64, a batch size of 16, and a gradient accumulation step of 16 to smooth out training. We fine-tuned all models for 3 epochs. We conducted fine-tuning on 4 Nvidia L40S GPUs.

\subsection{Prompting}

Our work explored various prompt engineering techniques to enhance the model's performance on the translation task. Accordingly, we tested on three distinct Prompting strategies.
\begin{enumerate}[left=0pt]

\item \textbf{Zero-shot Prompting: }
 According to researchers in \cite{Zhang2023PromptingLL}, an English template in a simple form works best for MT. Thus, we adopted the instruction prompt shown in Figure 2.

\item \textbf{Few-shot Prompting:} 
Additionally, we experimented with few-shot prompting, where examples of translations are provided. These examples can be randomly selected, however, research in \cite{Fernandes2023TheDI}\cite{Garca2023TheUE} shows that choosing good in-context examples can trigger the pre-trained language model to generate the desired output and also elicit the information learned during training. In addition, according to \cite{Fernandes2023TheDI}, the number and quality of prompt examples matter, where using suboptimal examples can potentially degenerate translation. We studied the best options in the ablation study in Section 4.1.  

\item \textbf{Contrastive Prompting:}
Besides the basic translation prompt, we opted to improve the quality of translations by guiding the model to generate the best translation from many options. To realize this goal, we extended the few-shot examples to include both good and bad translations\cite{Jiao2023ParroTTD}. This prompting technique mirrors fine-tuning on contrastive prompts. 

\end{enumerate}

\subsection{Evaluation}
\textbf{Evaluation Metric: }In the field of Neural Machine Translation (NMT), the accurate evaluation of translation quality remains a critical challenge. Altough traditional lexical-based metrics such as BLEU \cite{Papineni2002BleuAM} have been widely used, they often fall short in capturing the nuanced aspects of translation quality. In \cite{yakhni-chehab-2025-llms}, authors showed that reference-free xCOMET-10.7B model achieves the best correlation with human judgment, when it comes to translating Lebanese dialect. \newline\newline\textbf{Test Data:  } We used a subset of FLoRes dataset \citep{team2022NoLL}, a translated dataset from Wikinews developed as part of the NLLB project. We evaluated our models on 500 parallel Lebanese/English sentences from FloRes. \newline\newline \textcolor{red}{\textbf{Existing Evaluation Data do not capture the linguistic and cultural complexities of the Lebanese dialect. }} To ensure the authenticity and relevance of our evaluation data, we deliberately selected content that reflects the casual conversations and concerns of the Lebanese people, rather than relying on translated material. Our primary source was the \textsf{"Levantine Arabic Made Easier" }podcast\footnote{https://nasmaofny.libsyn.com/}, which offers a rich tapestry of bilingual stories from Lebanon. We identified around 15 episodes that were transcribed in Arabizi- a popular informal transliteration system used in electronic communication by Arabic speakers. Arabizi passages are then transformed into Arabic script using Yamli\footnote{https://www.yamli.com/arabic-keyboard/} platform, and then manually revised.
We used existing English translations of the episodes, which were produced by professional translators fluent in both Arabic and English. This provided us with 70 high-quality parallel data for our evaluation. We denote this dataset as \textbf{LebEval} (\textbf{Leb}anese \textbf{Eval}uation Dataset). 

\section{Results}
\subsection{Ablation Study}
\textbf{Number of Few-shot Examples: } To identify the optimal number of few-shot examples (K) for our model, we conducted experiments with three different settings: K=3, K=5, and K=7. \newline\newline\textbf{Selection of Few-shot Examples: } Apart from randomly selecting few-shot examples, we chose examples based on a certain criterion to increase their relevance to our input data. We used two distinct methods:

\begin{itemize}
\item \textbf{Embedding-based}: We generated embeddings for inputs and demonstrations using the LASER2 model. For each new input, we computed cosine similarities between its embedding and those of the example pool, selecting the top k most similar examples as demonstrations.
\item \textbf{Frequency-based Matching}: We identified examples containing rare bilingual expressions from the input text. Using a frequency matrix derived from a large Lebanese corpus, we specified bilingual words in our input sentence with a frequency below a certain threshold, and we selected examples containing these rare words. This approach prioritized examples containing challenging or uncommon Lebanese linguistic elements.\newline
\end{itemize} 
\textbf{Evaluation: } We evaluated the translation quality of few-shot prompting for Aya23-8B for three example selection methods: random, embedding-based, and frequency-based matching. We constructed the demonstrations' pool from the NN+LW sentences. Results are shown in Figure 4.\newline\newline \textbf{Results. } Our systematic evaluation revealed that K=3 achieved the best performance balance. Using 5 or 7 examples led to diminishing returns and increased computational overhead without significant performance gains. In addition, we show that selecting examples based on a criterion did not yield significant gains over random sampling while introducing significant overhead, especially in Matching-Based setting. In our main experiments, we opted for random sampling and we used k=3 for few-shot prompting and contrastive prompting denoted as \textbf{3-shot} and \textbf{C3-shot}, respectively.

\begin{figure}[h]
    \centering
    \includegraphics[width=1\linewidth]{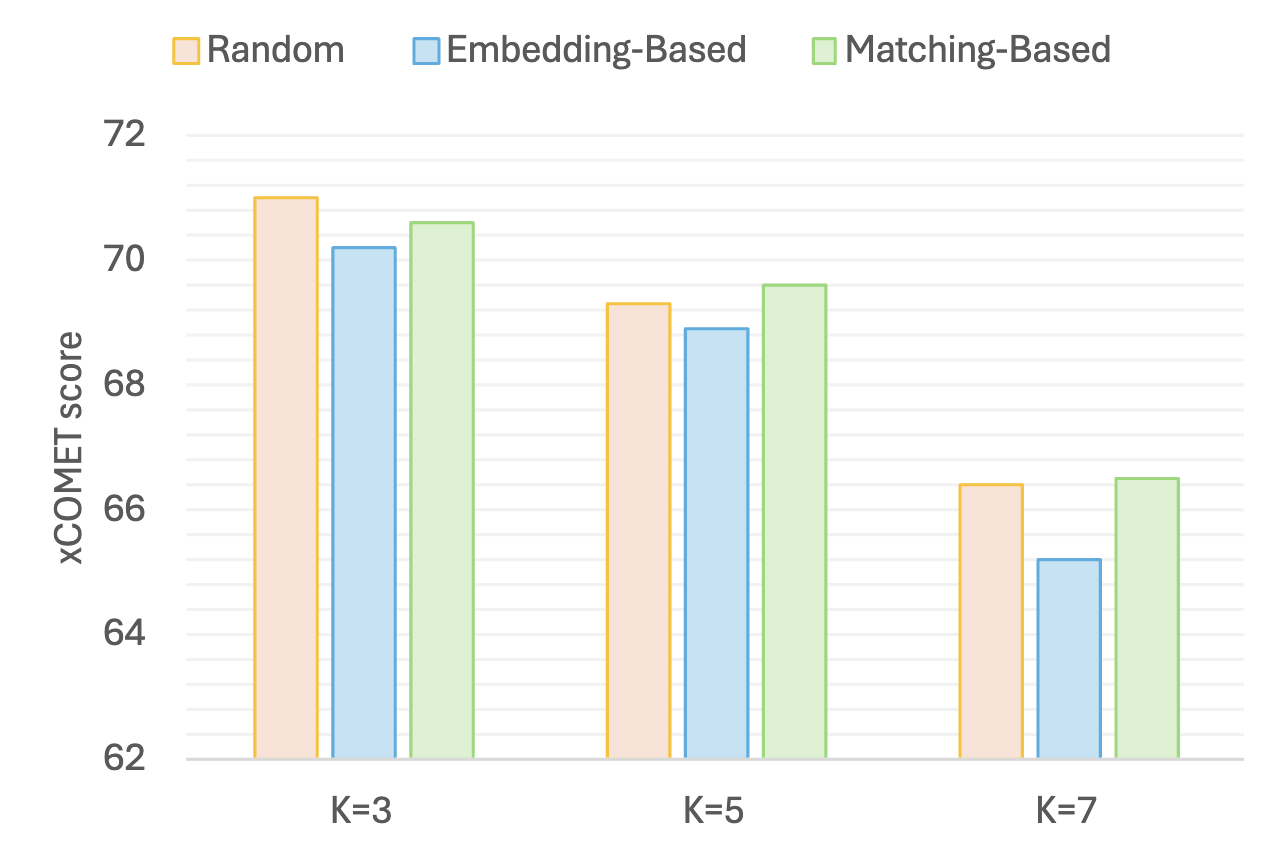}
    \caption{Impact of few-shot example selection methods (random, embedding-based, and frequency-based matching) and varying K values (K=3, 5, 7) on the translation quality of Aya23-8B.}
    \label{fig:enter-label}
\end{figure}

\subsection{Main Results}

\begin{table*}[h]
\centering
\caption{Translation performance of Cohere models on Flores subsets and our test set, for each of the configurations discussed in section 4.2. Best scores in each prompting setting are marked in \textbf{bold}. }
\begin{tabular}{p{0.2\columnwidth}p{0.55\columnwidth}ccclccc}
\hline
\multirow{2}{*}{Model} & \multirow{2}{*}{System}& \multicolumn{3}{c}{FLoRes} & &\multicolumn{3}{c}{LebEval} \\
\cline{3-5}\cline{7-9}
& & 0-shot & 3-shot & C3-shot & & 0-shot & 3-shot & C3-shot\\
\hline
\multirow{10}{*}{Aya23-8b} & \textit{Single-step Training} & & & & & &\\
& Vanilla & 85.5 & 87.2 & 87.5 & & 68.7& 71.0& 71.4\\
\cdashline{2-9}
& Instuct-MT-NN & \textbf{87.6} & 87.9 & 86.7 & & 70.9& 72.5& 71.1\\
& Instuct-MT-LW & 86.9 & 87.6 & 87.0 & & \textbf{73.6}& 72.9& 71.0\\
\cdashline{2-9}
& Instruct-Cont-NN & 87.2 & \textbf{88.3} & \textbf{89.1} & & 71.8& 72.8 & 73.2\\
& Instruct-Cont-LW & 86.8 & 87.4 & 87.4 & & 71.7& \textbf{73.5}& \textbf{74.4}\\
\cdashline{2-9}  
& Instruct-Gram & 84.1 & 86.1 & 86.4 & & 67.5 & 69.2 & 70.1\\
\cline{2-9}
& \textit{Curriculum Training} & & & & & &\\
& CONT+MT-NN & 87.0 & 88.2 & 88.7 & & 71.4& 72.5 & 72.4\\
& CONT+MT-LW & 86.9 & 87.3 & 87.5& & 71.4& 73.3& 74.1\\
\cdashline{2-9}
& Gram+CONT+MT-NN & 87.6 & 87.9 & 88.0 & & 72.0& 72.5 & 72.7\\
& Gram+CONT+MT-LW & 87.0 & 87.3 & 87.8 & & 72.0& 72.9 & 73.5\\
\hline

\end{tabular}
\end{table*}

\textbf{Fine-tuning on the Culturally-Aware Language Wave (LW) Dataset Consistently Yields Superior Results}: The xCOMET scores, reported in Table 1, demonstrate the effectiveness of adapters fine-tuned on LW. Across all prompting techniques, the results underscore the advantages of leveraging a culturally-aware native dataset for both standard and contrastive fine-tuning approaches. The superior performance emphasizes the advantage of Data Quality Over Quantity and highlights the critical role of culturally-rich datasets in accurately translating dialectal content. \newline\newline \textbf{Contrastive tuning outperforms basic translation, especially when coupled with contrastive prompting}: In both the few-shot and contrastive settings, contrastive fine-tuning delivered superior performance compared to basic instruction tuning, with Instruct-Cont-LW adapter achieving highest xCOMET score (74.4) on LebEval dataset. These findings underscore the advantages of integrating contrastive methods in both fine-tuning and prompting to help the model better address and understand translation errors.\newline\newline\textbf{Interestingly, curriculum learning did not result in notable performance improvements, regardless of the approach}. Specifically, neither teaching the model to learn from its mistakes before fine-tuning it on translation instructions \texttt{(CONT+MT)} nor introducing grammar rules as a preliminary step \texttt{(Grammar+CONT+MT)} yielded significant gains. This outcome may be attributed to the phenomenon of catastrophic forgetting, which is a well-documented limitation of large language models (LLMs).
A potentially more effective alternative could involve fine-tuning three separate models, each dedicated to one of these tasks: learning from errors, understanding grammar rules, and handling translation instructions. These specialized models could then be applied sequentially to leverage the strengths of curriculum learning across distinct stages. However, this approach was not pursued due to constraints in time and resources, as well as the additional complexity required to carefully redesign instruction formats. \newline\newline\textbf{ Culturally-relevant evaluation benchmarks are essential for accurately assessing model performance and addressing limitations in existing datasets.} Results demonstrate a clear advantage when models are evaluated on the native, culturally-aware LebEval dataset compared to FLORES. The base Aya model exhibits significantly better performance on FLORES than on LebEval, further emphasizing the disparity between culturally-generic benchmarks and datasets tailored to specific linguistic and cultural contexts. These findings underscore the need for more robust benchmarking efforts aimed at curating authentic evaluation data that accurately reflect the complexities of dialectal and culturally-rich language content.

\section{Preference Alignment vs. Fine-tuning}

Supervised Fine-Tuning (SFT) uses parallel datasets to train models to produce desired output, but may lack adaptability for cultural or stylistic nuances. In addition, SFT lacks a mechanism to prevent the model from rejecting mistakes in translations. To address these limitations, we investigated in Section 2.2 the use of contrastive instructions to guide the model in rejecting suboptimal translations. An alternative and potentially effective method is the use of preference-based techniques, such as Contrastive Preference Optimization (CPO) \citep{Xu2024ContrastivePO}. Preference alignment techniques use reinforcement learning to enable models to learn from ranked translations by prioritizing higher-quality outputs.

In this section, we investigate the effectiveness of CPO, a preference-based alignment method developed for translation tasks. To facilitate this, we constructed preference datasets as explained in Section 2.2, and fine-tuned the Aya23-8b model. The evaluation results on LebEval data, measured using xCOMET scores, are presented in Table 2.

\begin{table}[h]
\centering
\begin{tabular}{lcc}
 \toprule
 &  LebEval& FLoRes\\
 \hline
 Base&  68.7& 85.5\\
 CPO-NN&  63.7& 83.1\\
 CPO-LW&  67.1& 85.7\\
 Instruct-Cont-NN&  70.9& 87.6\\
 Instruct-Cont-LW&  73.6& 86.9\\
\toprule
\end{tabular}
\caption{Comparison of xCOMET scores for Aya23-8B fine-tuned with contrastive tuning (Instruct-Cont) and preference-based alignment (CPO) on LebEval and FLoRes test sets. }
\label{tab:my_label}
\end{table}

CPO consistently underperformed compared to standard SFT across all experimental configurations, often yielding results below the baseline model's performance. This persistent lower performance can be attributed to several factors, including the potential limitations of the preference data, as the rejected translations were sourced directly from the LLM itself. Additionally, the dialectal richness and cultural nuances of Lebanese Arabic introduce significant challenges for effective preference learning. A more systematic approach to preference data collection, focusing on a single aspect (such as cultural alignment) coupled with SFT on translation instructions, may yield more promising results. However, due to resource constraints in curating such specialized data, this approach was not explored in our study.

\section{Conclusion}
Our work demonstrates the critical importance of cultural authenticity in training LLMs for dialectal translation, particularly for Lebanese Arabic. Through extensive experiments with various instruction-tuning approaches and prompting strategies, we have shown that models trained on culturally-aware data consistently outperform those trained on larger but translated datasets. This finding challenges the common assumption that more training data necessarily leads to better performance, especially in the context of dialectal translation.

Furthermore, we show the advantage of using contrastive instruction tuning in translating dialectal Lebanese, which emphasizes the gained benefits of teaching the model to distinguish between good and poor translations. 

Finally, our introduction of LebEval as a culturally-aware evaluation benchmark has revealed substantial gaps between performance metrics on traditional benchmarks versus authentic dialectal content. This disparity underscores the importance of developing evaluation benchmarks that can effectively capture the nuances of dialectal translation.

\section{Limitations and Future Work}
This work presents some limitations. First, our experiments were constrained by the small size of culturally-aware datasets available for the Lebanese dialect, which limited our ability to fully explore the potential of various training approaches. Second, while our grammar-based instruction generation was based on Claude 3.5 Sonnet, the synthetic nature of these instructions may not fully capture the complexity of Lebanese grammatical structures. Additionally, our preference alignment experiments were limited by using model-generated rejected translations rather than human-curated examples, potentially affecting the quality of contrastive learning.

Our findings point to several promising research directions. Investigating efficient adaptation through the use of the mixture of experts (MoE) approach for MT tasks \citep{pham2023taskbasedmoemultitaskmultilingual} presents an intriguing avenue for LLM fine-tuning. Another promising approach in LLM fine-tuning for MT is the development of agentic models \citep{Barua2024ExploringAA} that improve grammatical, contrastive, and translation tasks. Additionally, building upon LebEval, research should aim to develop more comprehensive evaluation datasets, specifically aimed to capture dialectal nuances. We did not experiment with the largest Aya models from Cohere, due to computational resource constraints. However, examining this model could provide valuable insights into the efficacy of our proposed techniques. Additionally, it would be instructive to experiment with other recent open-source Arabic-centric LLMs such as Jais \citep{sengupta2023jais} and AceGPT \citep{huang-etal-2024-acegpt}. 

\bibliography{custom}

\onecolumn
\appendix
\newpage

\twocolumn
\section{Related Work}

\textbf{Prompting LLMs for MT}
Research by \citep{Hendy2023HowGA} and \citep{Jiao2023IsCA} shows that GPT models can perform translations effectively with appropriate prompting. However, they may face challenges with specialized content in certain language pairs, when compared to dedicated translation systems. Enhanced translation performance in open-source LLMs has been achieved through advanced prompting techniques, such as self-correction \citep{Feng2024ImprovingLM}, dictionary-based prompting \citep{Ghazvininejad2023DictionarybasedPP}, and mimicking human-like reasoning by breaking the translation process into smaller sub-tasks \citep{He2023ExploringHT}. Additionally, the use of autonomous agents within LLMs has been explored \citep{Barua2024ExploringAA}. Despite these innovations, translating low-resource languages remains a significant challenge. Notably, \citep{Tanzer2023ABF} and \citep{Zhang2024TeachingLL} highlight that LLMs can learn to translate new languages not present in their training data. This capability is further examined in a study leveraging LLMs for the translation of Saris, a low-resource language \citep{ondrejova-suppa-2024-llms}. \newline\newline\textbf{Finetuning LLMs for MT:} With the rise of powerful open-source LLMs such as BLOOM (cite appropriately) and LLaMA \citep{Touvron2023LLaMAOA}, there has been a surge in creating instruction-tuned models like Alpaca, Vicuna, and WizardLM \citep{Xu2024WizardLMEL}. While most efforts focus on general NLP tasks, recent work has emphasized fine-tuning LLMs for machine translation. Studies such as \citep{Li2023ElicitingTT} show that multilingual fine-tuning with explicit translation instructions significantly improves translation performance for diverse language pairs. Furthermore, fine-tuning using alignment instructions has shown consistent improvements in multiple translation directions, with error-guided alignments yielding further gains \citep{Mao2024TuningLW}. ParroT \citep{Jiao2023ParroTTD} is a framework that reformulates translation data into error-guided instructions to improve translation quality. 

Some strategies were explored for low-resource languages. Researchers in \citep{Zhang2023PLUGLP} developed PLUG, a framework that leverages Pivot languages to enhance instruction tuning for low-resource languages, while \citep{Iyer2023TowardsED} designed instruction datasets that address ambiguous sentences containing polysemous words and rare senses in an attempt to handle linguistic ambiguity of low-resource languages. Despite these advancements, the issue of translating low-resource languages remains largely unaddressed.\newline\newline\textbf{LLMs for Arabic MT:}
Limited work has focused on evaluating approaches to benchmark LLaMA3 for code-switched Arabic dialects \citep{Khondaker2024BenchmarkingLO}, while studies assessing commercial models such as ChatGPT and GPT-4 demonstrate their superiority over supervised baselines like NLLB in zero-shot settings \citep{kadaoui2023tarjamat}. Recent benchmarks, including LAraBench \citep{Abdelali2023LAraBenchBA} and SADID \citep{Abid2020TheSE}, have contributed to advancing Arabic machine translation (MT). However, SADID relies mainly on English-sourced translations, rather than authentic dialectal content, limiting its cultural and linguistic relevance. Despite these efforts, most research in Arabic MT has focused on benchmarking large language models (LLMs) rather than exploring their fine-tuning for Arabic dialect translation. The absence of studies dedicated to fine-tuning models specifically for this task highlights a critical gap in the field, underscoring the need for targeted approaches to improve translation quality for Arabic dialects. \newline\newline

\textbf{LLMs for Culturally-aware MT:} Despite dialects being deeply rooted in cultural context, the field continues to rely heavily on translated data.
Recent studies show that Large Language Models (LLMs) outperform traditional neural MT systems in handling cultural content and Culturally Specific Items (CSIs)\citep{yao-etal-2024-benchmarking}. While Arabic-centric LLMs like Jais \citep{sengupta2023jais} and AceGPT \citep{huang-etal-2024-acegpt} show promise, they face limitations due to their reliance on translated datasets. Although initiatives like Dallah \citep{Alwajih2024DallahAD} and evaluation benchmarks like AraDICE \citep{Mousi2024AraDiCEBF} have emerged, the challenge extends beyond isolated cultural items to the entire linguistic system. The field's continued dependence on translated data rather than authentic dialectal content indicates a pressing need for developing genuine, culturally aware datasets that fully capture Arabic dialectal variations.

\onecolumn
\newpage
\section{Grammar Data Synthesis}
\label{sec:appendix}

\begin{figure*}[h]
    \centering
    \includegraphics[width=1\linewidth]{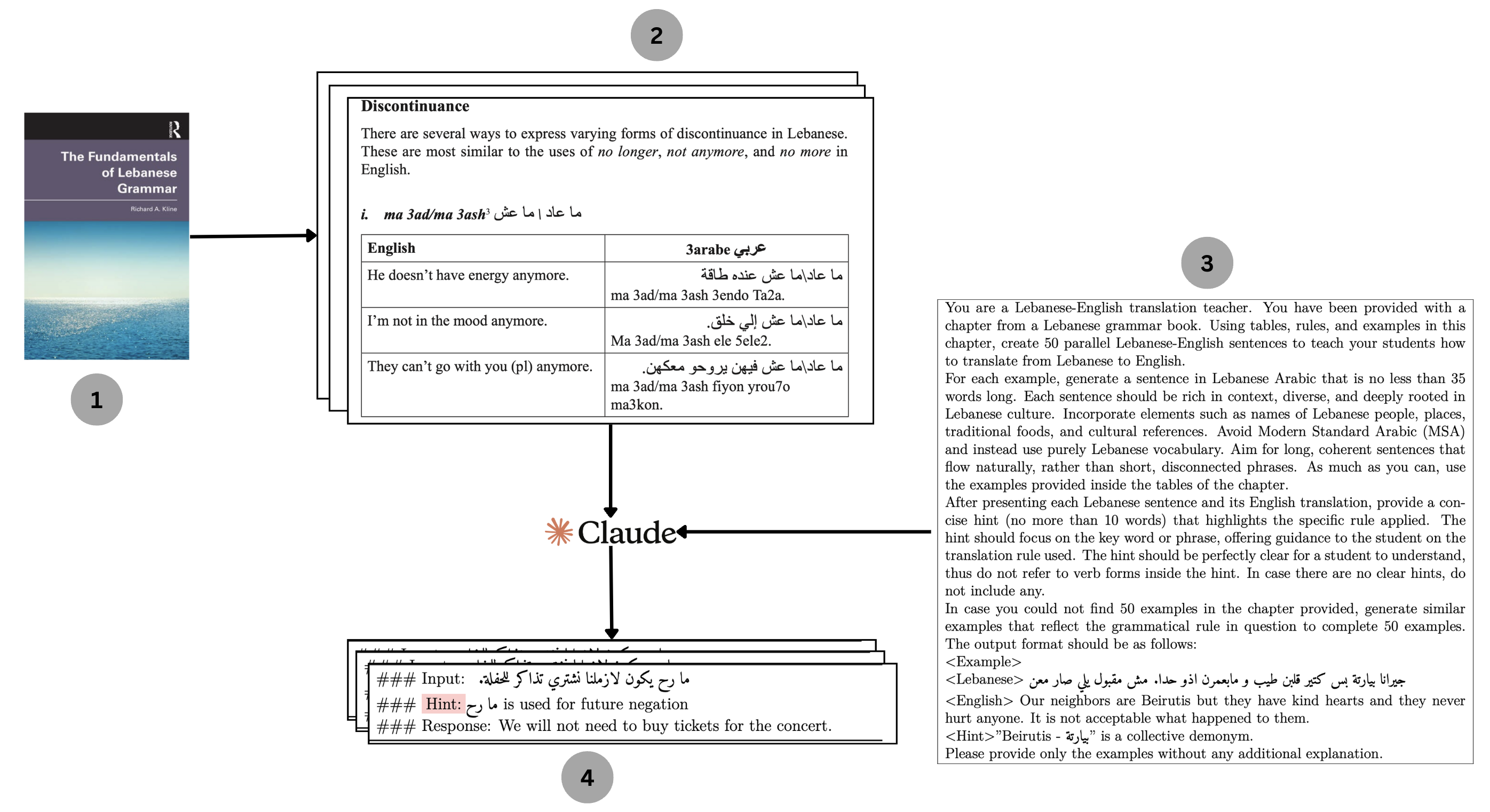}
    \caption{Steps performed to synthesize the Lebanese Grammatical Data: 1) Choosing a Lebanese Grammatical book, 2) Chunking the book into small Grammatical paragraphs, 3) Prompting Claude to use the small paragraphs, to 4) generate Grammatical instructions. }
    \label{fig:enter-label}
\end{figure*}
\newpage

\section{Qualitative Analysis}

\begin{figure*}[h]
    \centering
    \includegraphics[width=1\linewidth]{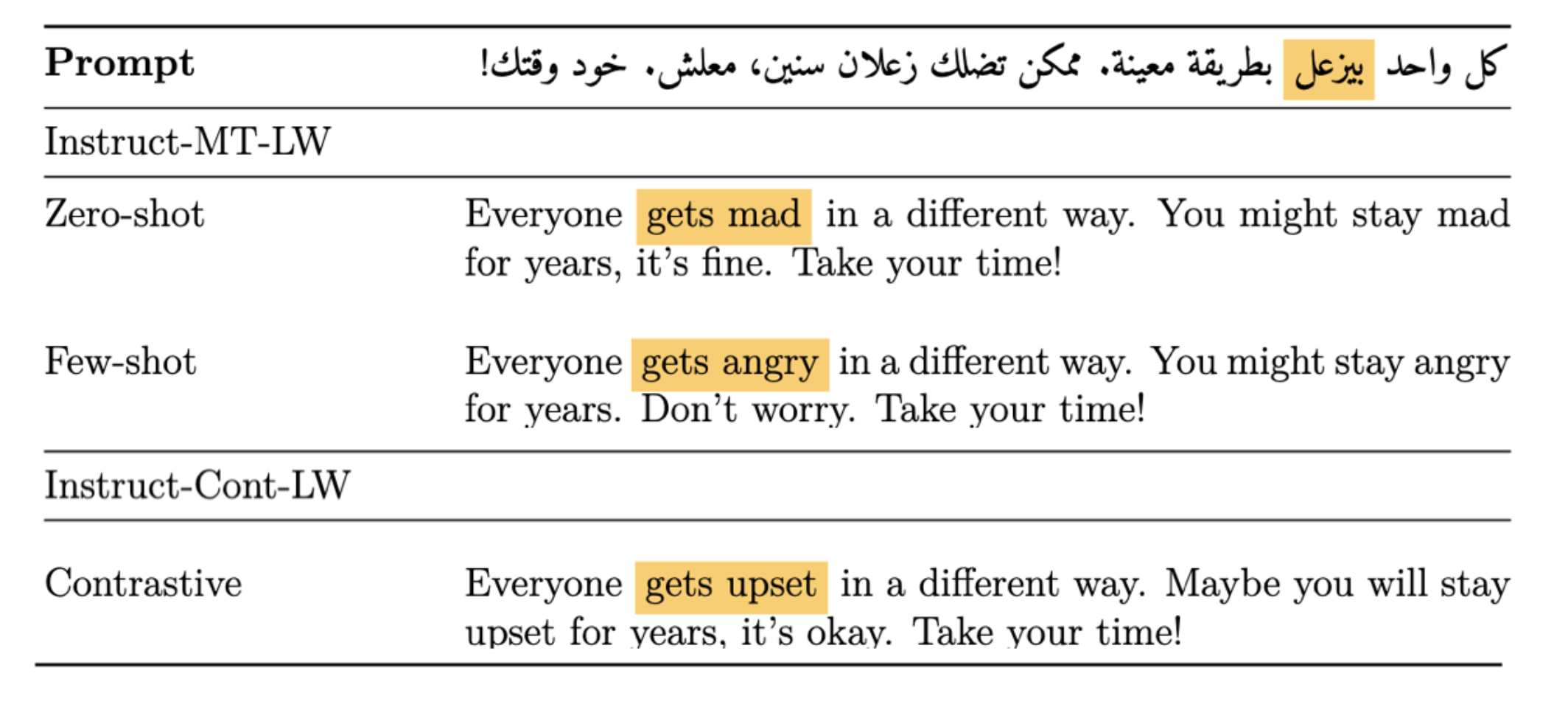}
    \label{fig:enter-label}
\end{figure*}

\begin{figure*}[h]
    \centering
    \includegraphics[width=1\linewidth]{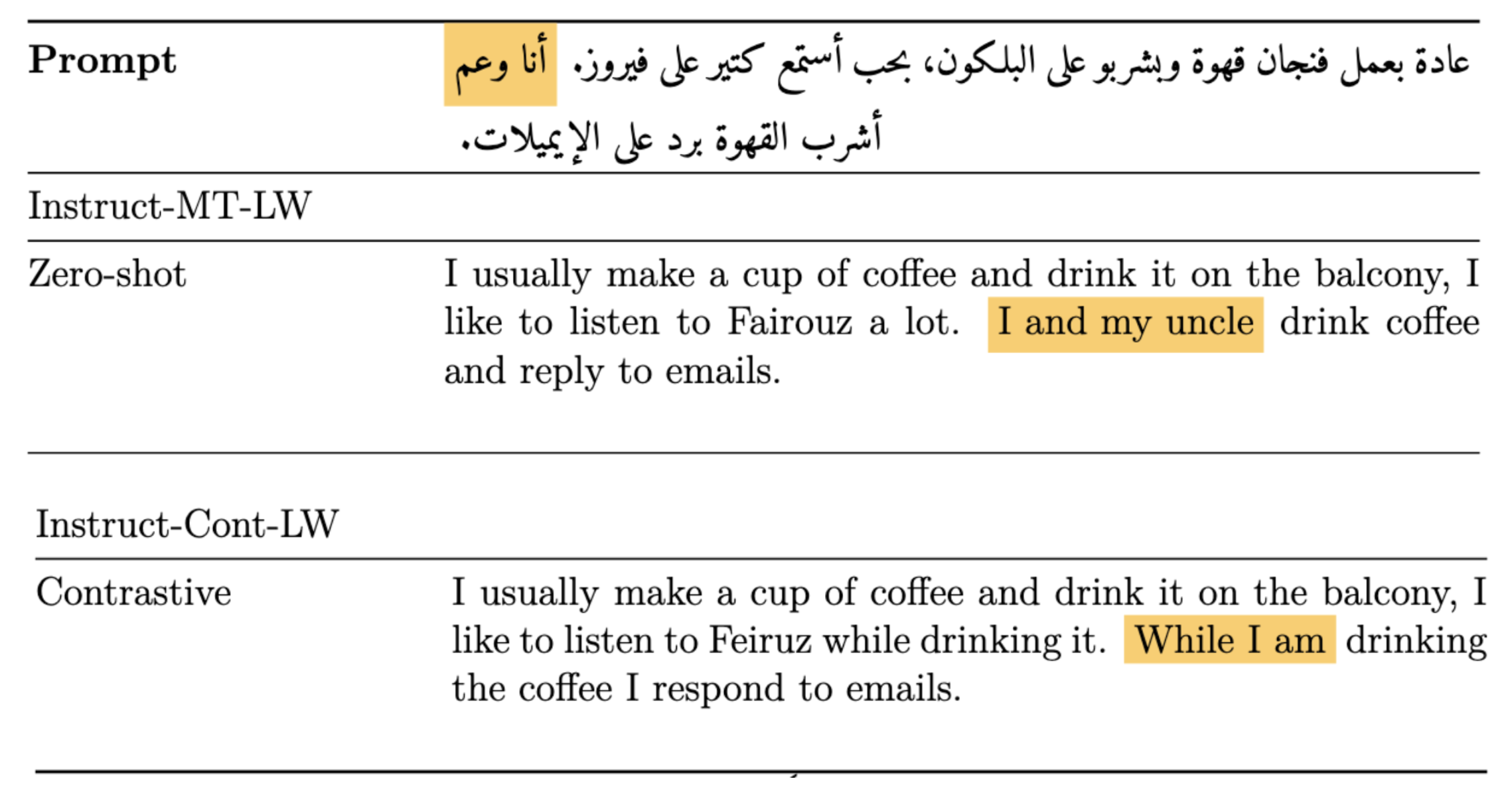}
    \caption{Qualitative Examples that show the superiority of adapters fine-tuned using Contrastive instructions. }
    \label{fig:enter-label}
\end{figure*}

\end{document}